\documentclass[11pt,a4paper]{article}
\usepackage[hyperref]{naaclhlt2019}
\usepackage{times}
\usepackage{latexsym}

\usepackage{comment}

\usepackage{amsmath}
\usepackage{cleveref}
\usepackage{booktabs}
\usepackage{multirow}
\usepackage{algorithm}
\usepackage[noend]{algorithmic}
\usepackage[normalem]{ulem}
\useunder{\uline}{\ul}{}
\usepackage{enumitem}
\usepackage{makecell}

\usepackage[backgroundcolor=White, textwidth=0.6in]{todonotes}

\aclfinalcopy 

\newcommand{\bfnew}{\\[5pt]}

\title{Analyzing the Perceived Severity of Cybersecurity Threats Reported on Social Media}

\author{Shi Zong$^1$, Alan Ritter$^1$, Graham Mueller$^2$ and Evan Wright$^3$\\
  $^1$The Ohio State University, OH, USA\\
  $^2$Leidos Inc., VA, USA\\
  $^3$FireEye LLC, CA, USA\\
  {\tt \{zong.56, ritter.1492\}@osu.edu}\\
  {\tt muellerwg@leidos.com}\\
  {\tt evan.wright@fireeye.com}\\}

\date{}

\begin{document}
\maketitle
\begin{abstract}
Breaking cybersecurity events are shared across a range of websites, including security blogs (FireEye, Kaspersky, etc.), in addition to social media platforms such as Facebook and Twitter.  In this paper, we investigate methods to analyze the severity of cybersecurity threats based on the language that is used to describe them online.  A corpus of 6,000 tweets describing software vulnerabilities is annotated with authors' opinions toward their severity.  We show that our corpus supports the development of automatic classifiers with high precision for this task.  Furthermore, we demonstrate the value of analyzing users' opinions about the severity of threats reported online as an early indicator of important software vulnerabilities.  We present a simple, yet effective method for linking software vulnerabilities reported in tweets to Common Vulnerabilities and Exposures (CVEs) in the National Vulnerability Database (NVD).  Using our predicted severity scores, we show that it is possible to achieve a Precision@50 of 0.86 when forecasting high severity vulnerabilities, significantly outperforming a baseline that is based on tweet volume.  Finally we show how reports of severe vulnerabilities online are predictive of real-world exploits.\footnote{Our code and data are available at {\tiny \urlstyle{sf} \url{https://github.com/viczong/cybersecurity_threat_severity_analysis}}.}
\end{abstract}


\section{Introduction}
Software vulnerabilities are flaws in computer systems that leave users open to attack; vulnerabilities are generally unknown at the time a piece of software is first published, but are gradually identified over time.  As new vulnerabilities are discovered and verified they are assigned CVE numbers (unique identifiers), and entered into the National Vulnerability Database (NVD).\footnote{\tiny \urlstyle{sf} \url{https://nvd.nist.gov/}}  To help prioritize response efforts, vulnerabilities in the NVD are assigned severity scores using the Common Vulnerability and Scoring System (CVSS).  As the rate of discovered vulnerabilities has increased in recent years,\footnote{\tiny \urlstyle{sf} \url{https://www.cvedetails.com/browse-by-date.php}} the need for efficient identification and prioritization has become more crucial.  However, it is well known that a large time delay exists between the time a vulnerability is first publicly disclosed to when it is published in the NVD; a recent study found that the median delay between the time a vulnerability is first reported online and the time it is published in the NVD is seven days; also, 75\% of threats are first disclosed online giving attackers time to exploit the vulnerability.\footnote{\tiny \urlstyle{sf} \url{https://www.recordedfuture.com/vulnerability-disclosure-delay/}}

\begin{figure}
    \centering
    \includegraphics[width=0.38\textwidth]{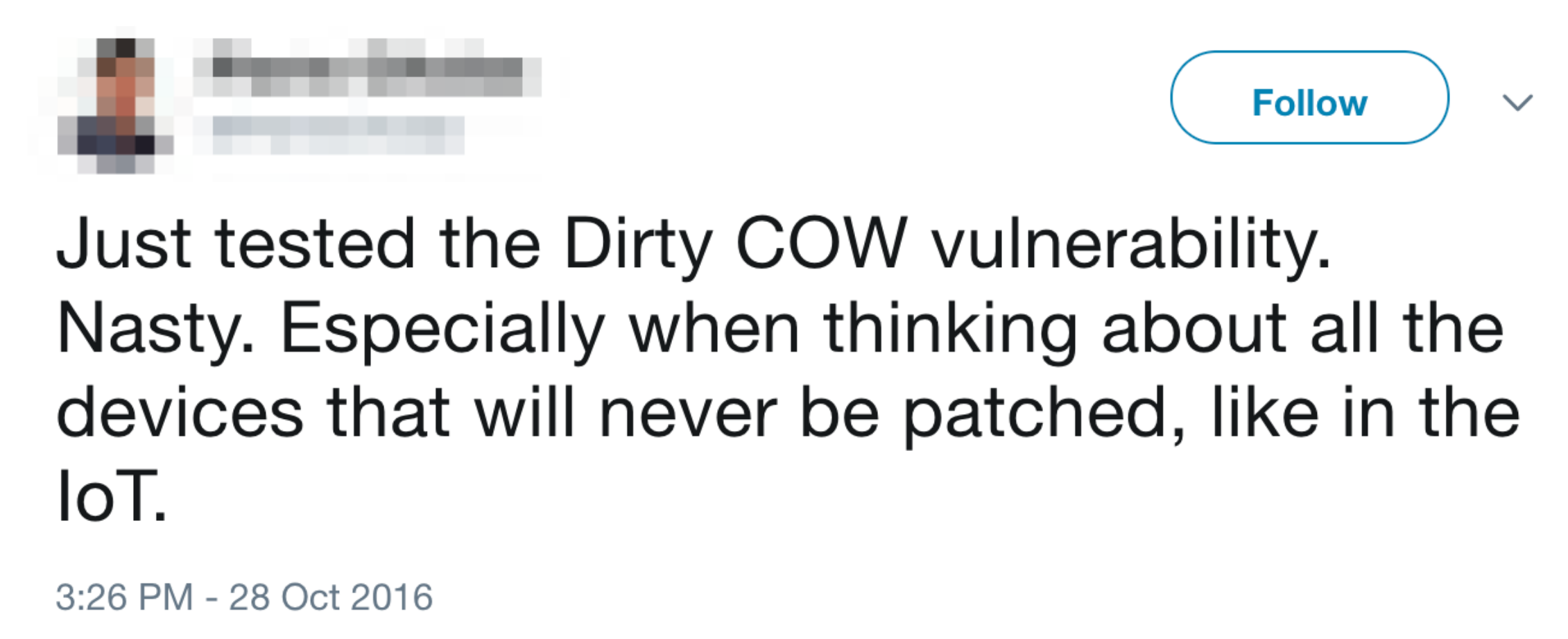}
    \caption{Example tweet discussing the dirty copy-on-write (COW) security vulnerability in the Linux kernel.}
    \label{fig:exampletweet1}
\end{figure}

In this paper we present the first study of whether natural language processing techniques can be used to analyze users' opinions about the severity of software vulnerabilities reported online.  We present a corpus of 6,000 tweets annotated with opinions toward threat severity, and empirically demonstrate that this dataset supports automatic classification.  Furthermore, we propose a simple, yet effective method for linking software vulnerabilities reported on Twitter to entries in the NVD, using CVEs found in linked URLs.  We then use our threat severity analyzer to conduct a large-scale study to validate the accuracy of users' opinions online against experts' severity ratings (CVSS scores) found in the NVD.  Finally, we show that our approach can provide an early indication of vulnerabilities that result in real exploits in the wild as measured by the existence of Symantec virus signatures associated with CVEs; we also show how our approach can be used to retrospectively identify Twitter accounts that provide reliable warnings about severe vulnerabilities.

Recently there has been increasing interest in developing NLP tools to identify cybersecurity events reported online, including denial of service attacks, data breaches and more \citep{ritter2015weakly,chang2016expectation,chambers2018detecting}.  Our proposed approach in this paper builds on this line of work by evaluating users {\em opinions} toward the severity of cybersecurity threats.

Prior work has also explored forecasting software vulnerabilities that will be exploited in the wild \citep{191006}.  Features included structured data sources (e.g., NVD), in addition to the volume of tweets mentioning a list of 31 keywords. Rather than relying on a fixed set of keywords, we analyze message content to determine whether the author believes a vulnerability is severe.
As discussed by \citet{191006}, methods that rely on tracking keywords and message volume are vulnerable to adversarial attacks from Twitter bots or sockpuppet accounts \citep{solorio2013case}.  In contrast, our method is somewhat less prone to such attacks; by extracting users' opinions expressed in individual tweets, we can track the provenance of information associated with our forecasts for display to an analyst, who can then determine whether or not they trust the source of information.


\section{Analyzing Users' Opinions Toward the Severity of Cybersecurity Threats}
\label{sec:data}

Given a tweet $t$ and named entity $e$, our goal is to predict whether or not there is a serious cybersecurity threat towards the entity based on context. For example, given the context in \Cref{fig:anno_interface}, we aim at predicting the severity level towards \textit{adobe flash player}.
We define an author's perceived severity toward a threat using three criteria: (1) does the author believe that their followers should be worried about the threat? (2) is the vulnerability easily exploitable? and (3) could the threat affect a large number of users?  If one or more of these criteria are met, then we consider the threat to be severe.

\begin{table*}[h!]
\centering
\resizebox{0.95\textwidth}{!}{
\begin{tabular}{c|l|c|c|l|c|c}
\toprule
\multirow{2}{*}{Anno. Tweets Total} & \multicolumn{3}{c|}{1st Annotation (5 workers per tweet)} & \multicolumn{3}{c}{2nd Annotation (10 workers per tweet)} \\ \cline{2-7}
 & Label & \# Tweets & \% & Label & \# Tweets & \%  \\\midrule
\multirow{3}{*}{6,000} & \multirow{2}{*}{With threat} & 2,543 & \multirow{2}{*}{42.4} & Severe threat & 506 & 25.7 \\ \cline{5-7} 
 &  &  (1,966 for 2nd anno.) &  & Moderate threat & 1,460 & 74.3 \\ \cline{2-7} 
 & Without threat & 3,457 & 57.6 & \multicolumn{3}{c}{/} \\
\bottomrule
\end{tabular}
}
 \caption{\label{tb:dataset_stats} Number of annotated tweets with break-down percentages to each category. In 1st annotation, a tweet contains a threat if more than 3 workers vote for it. In 2nd annotation, a threat is severe if more than 6 workers agree on it. Number of workers cut-offs are determined by comparing to our golden annotations in pilot studies.}
\end{table*}

\subsection{Data Collection}
\label{sec:data_collect}
To collect tweets describing cybersecurity events for annotation, we tracked the keywords ``ddos" and ``vulnerability" from Dec 2017 to July 2018 using the Twitter API.
We then used the Twitter tagging tool described by Ritter et. al. \shortcite{ritter-EtAl:2011:EMNLP} to extract named entities,\footnote{\tiny\urlstyle{sf}\url{https://github.com/aritter/twitter_nlp}} retaining tweets that contain at least one named entity.
To cover as many linguistic variations as possible, we used Jaccard similarity with a threshold of 0.7 to identify and remove duplicated tweets with same date.\footnote{We sampled a dataset of 6,000 tweets to annotate.}

\subsection{Mechanical Turk Annotation}
\label{sec:data_anno}
We paid crowd workers on Amazon Mechanical Turk to annotate our dataset.  The annotation was performed in two phases; during the first phase, we asked workers to determine whether or not the tweet describes a cybersecurity threat toward a target entity, in the second phase the task is to determine whether the author of the tweet believes the threat is severe; only tweets that were judged to express a threat were annotated in the second phase.  Each HIT contained 10 tweets to be annotated; workers were paid \$0.20 per HIT.  In pilot studies we tried combining these two annotations into a single task, but found low inter-rater agreement, especially for the threat severity judgments, motivating the need for separation of the annotation procedure into two tasks.

\begin{figure}[h!]
\centering
\includegraphics[width=0.49\textwidth]{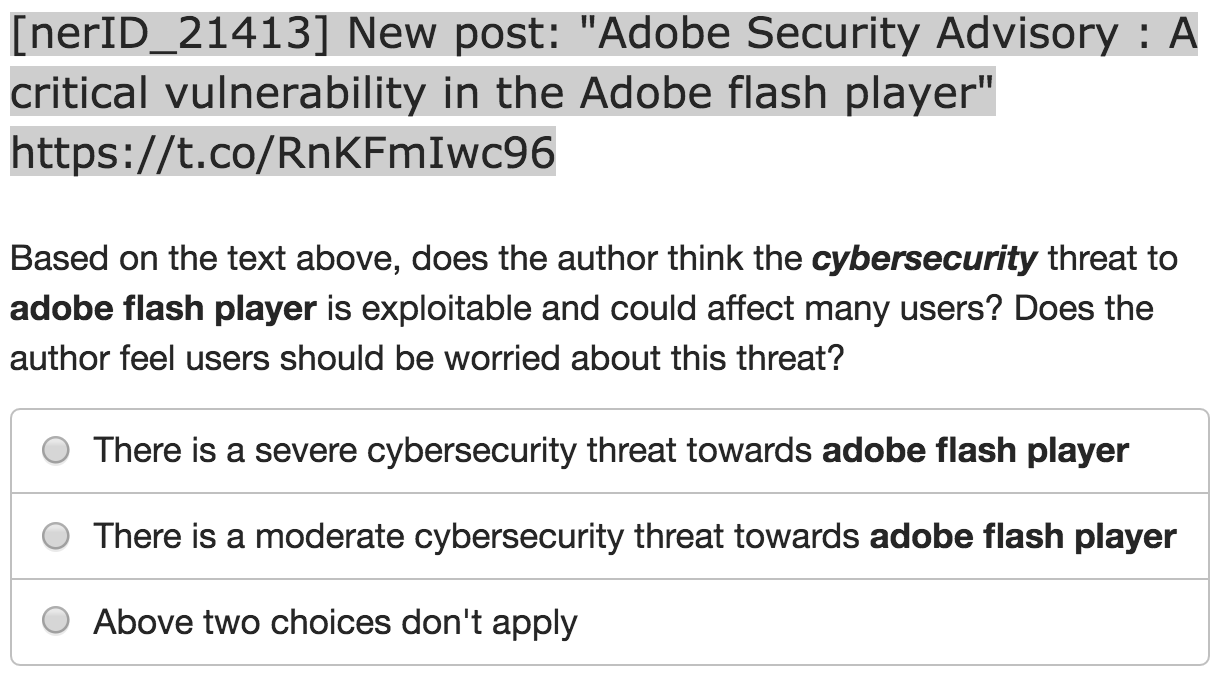}
\caption{A portion of the annotation interface shown to MTurk workers during the threat severity annotation.}
\label{fig:anno_interface}
\end{figure}

\Cref{fig:anno_interface} shows a portion of the annotation interface presented to workers during the second phase of annotation.  Details of each phase are described below, and summarized in  \Cref{tb:dataset_stats}.
\bfnew
\textbf{Threat existence annotation:} Not all tweets in our dataset describe cybersecurity threats, for example many tweets discuss different senses of the word ``vulnerability" (e.g., ``It's OK to show vulnerability").
During the first phase of our annotation process, workers judged whether or not there appears to be a cybersecurity threat towards the target entity based on the content of the corresponding tweet.  We provide workers with 3 options: the tweet indicates (a) a cybersecurity threat towards given entity, (b) a threat, but not towards the target entity, or (c) no cybersecurity threat. Each tweet is annotated by 5 workers.
\bfnew
\textbf{Threat severity annotation:} In the second phase, we collect all tweets judged to contain threats by more than 3 workers in the first phase and annotated them for severity. 1,966 tweets were selected out of 6,000.\footnote{We further deduplicate pairs of tweets where the longest common subsequence covers the majority of the text contents.  During deduplication all hashtags and URLs were removed and digits were replaced with 0.} For each tweet we provided workers with 3 options: the tweet contains (a) a severe, (b) a moderate or (c) no threat toward the target entity.  During our pilot study, we found this to be a more challenging annotation task, therefore we  increased the number of annotators per tweet to 10 workers, which we found to improve agreement with our expert judgments.
\bfnew
\textbf{Inter-annotator agreement:} During both phases, we monitored the quality of workers' annotations using their agreement with each other. We calculated the annotation agreement of each worker against the majority vote of other workers.
We manually removed data from workers who have an agreement less than 0.5, filling in missing annotations with new workers.
We also manually removed data from workers who answered either uniformly or randomly for all HITs.
\bfnew
\textbf{Agreement with expert judgments:} To validate the quality of our annotated corpus we compared the workers' aggregated annotations against our own expert annotations. We independently annotated 150 randomly sampled tweets, 61 tweets of which are marked as containing severe or moderate threats.  For threat existence annotation, we observe a 0.66 value of Cohen's $\kappa$ \cite{doi:10.1162_coli.07-034-R2} between the expert judgements and majority vote of 5 crowd workers. Although our threat severity annotation task may require some cybersecurity knowledge for accurate judgment, we still achieve 0.52 Cohen $\kappa$ agreement by comparing majority vote from 10 workers with expert annotations.


\subsection{Analyzing Perceived Threat Severity}
\label{sec:predict}

Using the annotated corpus described in \Cref{sec:data_anno}, we now develop classifiers that detect threats reported online and analyze users' opinions toward their severity.
Specifically, given a named entity and tweet, $\langle e, t\rangle$, our goal is to estimate the probability the tweet describes a cybersecurity threat towards the entity, $p_{\text{threat}}(y|\langle e, t\rangle)$ and also the probability that the threat is severe, $p_{\text{severe}}(y|\langle e, t\rangle)$. In this section, we describe the details of these classifiers and evaluate their performance.


We experimented with two baselines to detect reports of cyberthreats and analyze opinions about their severity: logistic regression using bag-of-ngram features, and 1D convolutional neural networks.  In the sections below we describe the input representations and details of these two models.
\bfnew
\noindent
{\bf Logistic regression:} 
We use logistic regression as our first baseline model for both classifiers. Input representations are bag-of-ngram features extracted from the entire tweet content.  Example features are presented in \Cref{tb:lr_top_ranked_features}.
We use context windows of size 2, 3 and 4 to extract features.  We map extracted $n$-grams that occur only once to a $\langle${\ttfamily UNK}$\rangle$ token.
In all our experiments, we replace named entities with a special token $\langle${\ttfamily TARGET}$\rangle$;
this helps prevent our models from biasing towards specific entities that appear in our training corpus. All digits are replaced with 0. 
\bfnew
\noindent
{\bf Convolutional neural networks:} 
We also experimented with 1D convolutional neural networks \cite{JMLR:v12:Ronan2011, kim:2014:EMNLP2014}. Given a tweet, the model first applies convolutional operations on input sequences with various filters of different sizes. The intermediate representations for each filter are aggregated using max-pooling over time, followed by a fully connected layer.
We choose convolution kernel sizes to be 3, 4 and 5-grams with 100 filters for each.  We minimize cross-entropy loss using Adam \cite{iclr:adam}; the learning rate is set to 0.001 with a batch size of 1 and 5 epochs.
\bfnew
\textbf{Word embeddings:} We train our own cybersecurity domain word embeddings based on GloVe \cite{pennington2014glove}, as 39.7\% of our tokens are treated as OOV words in GloVe pre-trained Twitter embeddings.  We used a corpus of 609,470 cybersecurity-related tweets (described in \Cref{sec:data_collect}) as our training corpus. The dimension of word embeddings is 50. \Cref{tb:cyber_embed_words} shows nearest neighbors for some sampled cybersecurity terms based on the learned embeddings.

During network training, we initialize word embedding layer with our own embeddings. We initialize tokens not in our trained embeddings by randomized vectors with uniform distribution from -0.01 to 0.01.  We fine-tune the word embedding layer during training.

\begin{table}[h!]
\begin{center}
\resizebox{0.48\textwidth}{!}{%
\begin{tabular}{l|p{0.4\textwidth}}
\toprule
Token & Nearest Neighbors\\\midrule
\#ddos & attacks, ddos, datacenter-insider, attack, \#cyberattack\\\hline
\#hackers & hackers, \@sec\_cyber, \#blackberryz00, \#malware, \#hacking\\\hline
threats & defenses, cyberrisk, \#cybersecurity, threat, \#iot-based \\\hline
vulnerability & risk, ..., \#vulnerability, strength, critical \\
\bottomrule
\end{tabular}
}
\end{center}
\caption{\label{tb:cyber_embed_words} Nearest neighbors to some cybersecurity related tokens in our trained word embeddings. Embeddings are trained by using GloVe. Similar tokens are sorted by cosine similarity scores.}
\end{table}

\subsubsection{Experimental Setup}

For threat existence classification, we randomly split our dataset of 6,000 tweets into a training set of 4,000 tweets, a development set of 1,000 tweets, and test set of 1,000 tweets. 
For the threat severity classifier, we only used data from 2nd phase of annotation.  This dataset consists of 1,966 tweets that were judged by the mechanical turk workers to describe a cybersecurity threat towards the target entity.  We randomly split this dataset into a training set of 1,200 tweets, a development set of 300 tweets, and a test set of 466 tweets. We collapsed the three annotated labels into two categories based on whether or not the author expresses an opinion that the threat towards the target entity is severe.

\subsubsection{Results}
\begin{figure}[h!]
\centering
\includegraphics[width=0.5\textwidth]{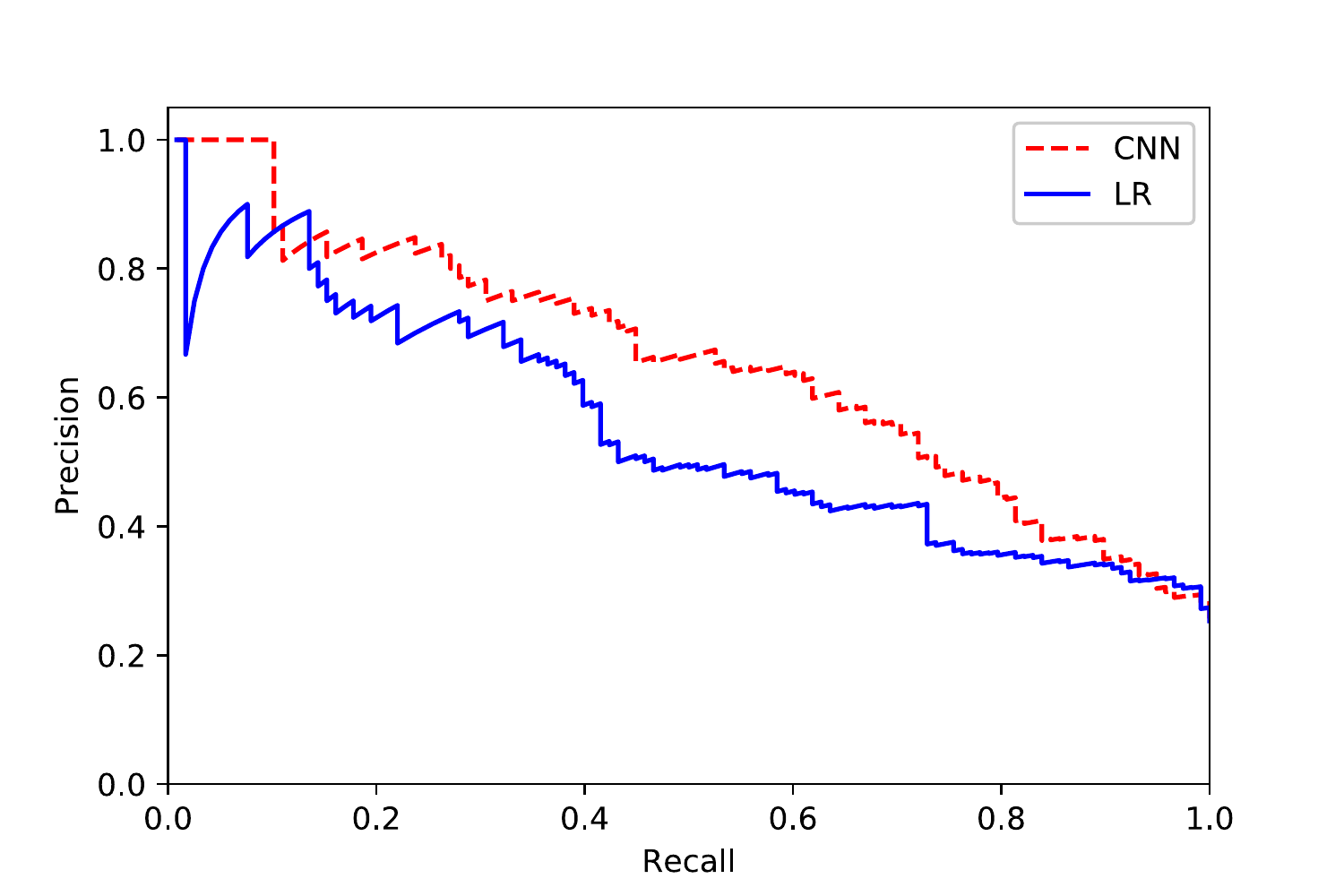}
\caption{Precision/Recall curves showing performances of convolutional model (CNN) and logistic regression model (LR) for threat severity classification task in test set.}
\label{fig:severe_pr}
\end{figure}

{\bf Threat existence classifier:}
The logistic regression baseline has good performance at identifying threats, which we found to be a relatively easy task; area under the precision-recall curve (AUC) on the development and test set presented in Table \ref{tb:model_perf_auc}.  This enables accurate detection of trending threats online by tracking cybersecurity keywords using the Twitter streaming API, following an approach that is similar to prior work on entity-based Twitter event detection \citep{ritter2012open,zhou2014simple,wei2015bayesian}.  \Cref{tb:threat_exist} presents an example of threats detected using this procedure on Nov. 22, 2018.\footnote{A live demo is available at: \tiny \urlstyle{sf}\url{http://kb1.cse.ohio-state.edu:8123/events/threat}}
\bfnew
{\bf Threat severity classifier:}
\Cref{fig:severe_pr} shows precision recall curves for the threat severity classifiers. Logistic regression with bag-of-ngram features provides a strong baseline for this task. \Cref{tb:lr_top_ranked_features} presents examples of high-weight features from the logistic regression model.  These features often intuitively indicate severe threats, e.g. ``critical vulnerability", ``a massive", ``million", etc.  Without much hyperparameter tuning on the development set, the convolutional neural network consistently achieves higher precision at the same level of recall as compared to logistic regression.  We summarize the performance of our threat existence and severity classifiers in \Cref{tb:model_perf_auc}.

\begin{table*}[!h]
\centering
\small
\begin{tabular}{l|p{0.64\textwidth}cc}
\toprule
Named Entity & Example Tweet & Existence & Severity \\\midrule
apple &	RT \@AsturSec: A kernel vulnerability in Apple devices gives access to remote code execution - Packt Hub \#infosec \#CyberSecurity https://t.... & 0.96 & 0.59\\
google & RT \@binitamshah: Unfixed spoofing vulnerability in Google Inbox mobile apps https://t.co/TWx7jSi1gc & 0.78 & 0.17 \\
adobe &	RT \@Anomali: Adobe released patches for three ``important-ranked'' severity vulnerabilities, including one vulnerability in Adobe Acrobat and... & 0.76 & 0.32 \\
flash & Vulnerability in Flash player allowing code execution. Patch before Black Friday: https://t.co/4idb570d1E \#CyberSecurity \#vulnerability & 0.71 & 0.43 \\
mac & adobe's flash player for windows, mac and linux has a critical vulnerability that should be patched as a top priori... https://t.co/LLlPATy9vR & 0.69 & 0.88\\
\bottomrule
\end{tabular}
\caption{Top five threats extracted with highest confidence on Nov. 22, 2018.  For each entity we aggregate tweets, and average threat existence scores.  The tweet with the maximum threat severity score is shown in each instance.}
\label{tb:threat_exist}
\end{table*}

\begin{table}[h!]
\centering
\small
\resizebox{0.48\textwidth}{!}{%
\begin{tabular}{lc|lc}
\toprule
Features & Weight & Features & Weight\\\midrule
ddos attack & 1.40 & $\langle$TARGET$\rangle$ , & 0.89\\
hackers to & 1.11 & take over & 0.87 \\
a massive & 1.07 & 00 countries & 0.85\\
critical vulnerability & 1.03 & attackers to & 0.84\\
0 billion & 0.96 & discovered in & 0.82\\
lets attackers & 0.95 & 000 million & 0.82\\
$\langle$TARGET$\rangle$ users & 0.91 & : \#ddos & 0.81\\
a critical & 0.91 & abuse and & 0.81\\
of a &  0.89 & , ddos & 0.81 \\
many $\langle$TARGET$\rangle$ & 0.89 & a severe & 0.79 \\
\bottomrule
\end{tabular}}
\caption{High-weight $n$-gram features from logistic regression model for threat severity classification task.}
\label{tb:lr_top_ranked_features}
\end{table}

\begin{table}[h!]
\centering
\small
\begin{tabular}{ll|cc}
\toprule
Task & Model & Dev AUC & Test AUC\\\midrule
Existence & LR & 0.88 & 0.85 \\\midrule
\multirow{2}{*}{Severity} & LR & 0.62 & 0.54 \\
& CNN & 0.70 & 0.65 \\
\bottomrule
\end{tabular}
\caption{Performance of our threat existence and severity classifiers. We show area under the precision-recall curve (AUC) for both development and test sets.}
\label{tb:model_perf_auc}
\end{table}


\section{Forecasting Severe Cybersecurity Threats}
\label{sec:forecast}

In Section \ref{sec:data} we presented methods that can accurately detect threats reported online and analyze users' opinions about their severity.  
We now explore the effectiveness of this model for forecasting. Specifically, we aim to answer the following questions:
{\bf (1)} To what extent do users' opinions about threat severity expressed online align with expert judgments?
{\bf (2)} Can these opinions provide an early indicator to help prioritize threats based on their severity?

\noindent {\bf A large corpus of users' opinions:}
We follow the same procedure described in \Cref{sec:data_collect} to prepare another dataset for a large-scale evaluation.  For this purpose, we collected data from Jan 2016 to Nov 2017; this ensures no tweets overlap with those that were annotated in \Cref{sec:data_anno}. We collect all English tweets that explicitly contain the keyword ``vulnerability" within this time period, which results in a total number of 976,180 tweets. 377,468 tweets remain after removing tweets without named entities.
\bfnew
\noindent
{\bf National Vulnerability Database (NVD):} 
NVD is the U.S. government database of software vulnerabilities.  Started in 2000, NVD covers over 100,000 vulnerabilities, assigning a unique CVE number for each threat. These CVE numbers serve as common identifiers.
NVD uses the Common Vulnerability Scoring System (CVSS) to measure the severity of threats. 
CVSS currently has two versions: CVSS v2.0 and CVSS v3.0 standards. CVSS v3.0 is the latest version released in July 2015. We summarize the two standards in \Cref{tb:cvss}.\footnote{\tiny \urlstyle{sf} \url{https://nvd.nist.gov/vuln-metrics/cvss}}

\begin{table}[h!]
\small
\begin{center}
\begin{tabular}{lc|lc}
\toprule
Severity & Base Score & Severity & Base Score\\\midrule
&& None & 0.0\\
Low & 0.0-3.9 & Low & 0.1-3.9\\
Medium & 4.0-6.9 & Medium & 4.0-6.9\\
High & 7.0-10.0 & High & 7.0-8.9\\
&& Critical & 9.0-10.0\\
\bottomrule
\end{tabular}
\end{center}
\caption{Qualitative severity rankings of vulnerabilities in NVD. (Left) CVSS v2.0 standards and (Right) CVSS v3.0 standards.}
\label{tb:cvss}
\end{table}

\noindent
{\bf Matching tweets with NVD records:} 
Evaluating our forecasts of high severity vulnerabilities relies on accurately matching tweets describing vulnerabilities to their associated NVD records.  To achieve this we present a simple, yet effective method that makes use of content in linked webpages.
We find that 82.4\% of tweets contain external urls in our dataset. 

Our approach to link tweets to CVEs is to search for CVE numbers either in url addresses or in corresponding web pages linked in tweets reporting vulnerabilities.\footnote{\label{fn:working_mechanism} Readers may be wondering why a CVE number has been generated before it is officially published in the database.  This is due to the mechanism of assigning CVEs. Some identified companies have the right to assign CVEs or have already reversed some CVEs. When a threat appears, a CVE number is assigned immediately before any further evaluation. NVD only officially publishes a threat after all evaluations are completed. Therefore, there is a time delay between CVE entry established date and the official publication date.} We ignore web pages that contain more than one unique CVE to avoid potential ambiguities.  Using this approach, within our dataset, 79,383 tweets were linked to 10,565 unique CVEs. In order to stimulate a forecasting scenario, we only consider CVEs where more than two associated tweets were posted at least 5 days ahead of official NVD publication date. In our dataset, 13,942 tweets are finally selected for forecast evaluation, covering 1,409 unique CVE numbers. 
To evaluate the accuracy of this linking procedure, we randomly sampled 100 matched pairs and manually checked them. We find the precision of our matching procedure to be very high: only 2 mismatches out of 100 are found.

\subsection{Forecasting Models}

Now that we have a linking between tweets and CVE numbers, our goal is to produce a sorted list of CVEs with those that are indicated to be severe threats the top.  We consider two ranking procedures, detailed below; the first is based on users' opinions toward the severity of a threat, and the second is a baseline that simply uses the volume of tweets describing a specific vulnerability to measure its severity.  To simplify the exposition below, we denote each CVE number as $\text{CVE}_i$, and the collection of tweets linked to this CVE number as $T_{\text{CVE}_i} = \{k|\text{tweet } t_k \text{ is mapped to } \text{CVE}_i\}$.
\bfnew
\textbf{Our model:} Our severe threat classifier assigns a severity score $p_{\text{severity}}(y|\langle e, t\rangle)$ for each tuple of name entity $e$ and corresponding tweet $t$. For a specific CVE, we define our severity forecast score to be the maximum severity scores among all tuples from matched tweets $\langle \cdot, t_k\rangle$ (a single tweet may contain more than one name entity):
\begin{align*}
(\text{CVE}_i)_{\text{forecast score}} = \max_{k \in T_{\text{CVE}_i}}p_{\text{severity}}(y|\langle \cdot, t_k\rangle).
\end{align*}

\noindent \textbf{Tweet volume baseline:} Intuitively, the number of tweets and retweets can indicate people's concern about a specific event. 
Specifically, the severity for threat $\text{CVE}_i$ according to the volume model is defined by the cardinality of $T_{\text{CVE}_i}$:
\begin{align*}
(\text{CVE}_i)_{\text{volume score}} = |T_{\text{CVE}_i}|.
\end{align*}


\subsection{Forecasting CVSS Ratings}
\label{sec:fore_cvss}
In our first set of experiments, we compare our forecasted threat severity scores against CVSS ratings from the NVD.
We define a threat as being severe if its CVSS score is $\geq$ 7.0. This cut-off corresponds to qualitative severity ratings provided by CVSS (marked as HIGH or CRITICAL in \Cref{tb:cvss}).\footnote{The Forum of Incident Response and Security Teams (FIRST) also provides an example guideline that recommends patching all vulnerabilities with CVSS scores $\geq$ 7.0. See {\tiny \urlstyle{sf}\url{https://www.first.org/cvss/cvss-based-patch-policy.pdf}}.}  We use the newest v3.0 scoring system, which was developed to improve v2.0.\footnote{\tiny \urlstyle{sf}\url{https://www.first.org/cvss/user-guide}} Large software vendors have announced of the adaptation of the CVSS v3.0 standards, including Cisco, Oracle, SUSE Linux, and RedHat.

We evaluate our models' performance at identifying severe threats five days ahead of the NVD publication date, within their top $k$ predictions.
\Cref{tb:forecast_prec} shows our results.  We observe that tweet volume performs better than a random baseline; having a large number of tweets beforehand is a good indicator for high severity, however our approach which analyzes the content of messages discussing software vulnerabilities achieves significantly better performance; 86\% of its top 50 forecasts were indeed rated as HIGH or CRITICAL severity in the NVD.

\begin{table}[!htbp]
\resizebox{0.48\textwidth}{!}{%
\centering
\begin{tabular}{l|cccc}
\toprule
 & P@10 & P@50 & P@100 & AUC\\\midrule
Random & 59.0 & 61.2 & 58.8 & 0.595\\
Volume model & 70.0 & 68.0 & 70.0 & 0.583\\
Our model & 100.0 & 86.0 & 78.0 & 0.658\\
\bottomrule
\end{tabular}
}
\caption{\label{tb:forecast_prec} Model performance of identifying severe threats (CVSS scores $\geq$ 7.0) with Precision@$k$ and area under the precision-recall curve (AUC) metrics. For majority random baseline, we average over 10 trails.}
\end{table}

\begin{table*}[!h]
\renewcommand\arraystretch{1.2}
\centering
\small
\resizebox{\textwidth}{!}{%
\begin{tabular}{c|p{0.15\textwidth}p{0.65\textwidth}p{0.16\textwidth}p{0.15\textwidth}}
\toprule
 & \begin{tabular}[t]{@{}l@{}} CVE Num /\\ Name Entity\end{tabular} & CVE Description / Matched Tweets & \begin{tabular}[t]{@{}l@{}} CVSS Scores / \\ Our Severity\end{tabular} & \begin{tabular}[t]{@{}l@{}} Publish Date\\ (\# Days Ahead)\end{tabular} \\ \toprule
 
 \multirow{4}{*}{(a)} & CVE-2016-0728 & The join\_session\_keyring function in security/keys/process\_keys.c in the Linux kernel before 4.4.1 mishandles object references in a certain error case, which allows local users to gain privileges or cause a denial of service (integer overflow and use-after-free) via crafted keyctl commands.
& \begin{tabular}[t]{@{}l@{}}7.2 HIGH (v2.0) \\ 7.8 HIGH (v3.0)\end{tabular} & 2016-02-08 \\\cmidrule{2-5}
 & Android & Vulnerability in the Linux kernel could allow attackers to gain access to millions of Android devices! http://thenextweb.com/insider/2016/01/20/newly-discovered-security-flaw-could-let-hackers-control-66-of-all-android-devices/ ... & 0.98
& 2016-01-20 (+19) \\
 & Android & A Serious Vulnerability in the Linux Kernel Hits Millions of PCs, Servers and Android Devices http://ift.tt/1OvB4JA &  0.89 & 2016-01-20 (+19) \\
 & Android & Millions of PCs and Android devices are at risk from a recently discovered critical zero-day vulnerability. http://goo.gl/r95ZYZ  \#infosec & 0.89
 & 2016-01-20 (+19) \\\toprule
 
\multirow{4}{*}{(b)} & CVE-2017-6753 & A vulnerability in Cisco WebEx browser extensions for Google Chrome and Mozilla Firefox could allow an unauthenticated, remote attacker to execute arbitrary code with the privileges of the affected browser on an affected system. 
& \begin{tabular}[t]{@{}l@{}}9.3 HIGH (v2.0) \\ 8.8 HIGH (v3.0)\end{tabular} & 2017-07-25 \\\cmidrule{2-5}
 & Cisco WebEx Extensions & The Hacker News : Critical RCE Vulnerability Found in Cisco WebEx Extensions, Again - Patch Now! http://ow.ly/gR3l30dJXlj  \#CDTTweets & 0.98 & 2017-07-19 (+6)\\
 & Cisco Systems & A critical vulnerability has been discovered in the Cisco Systems' WebEx browser extension for \#Chrome and \#Firefox: http://s.cgvpn.net/Zu  & 0.94 & 2017-07-18 (+7) \\
 & Cisco WebEx Extensions & ``Critical RCE Vulnerability Found in Cisco WebEx Extensions, Again - Patch Now!" via The Hacker News \#security http://ift.tt/2va8Wrx & 0.93 & 2017-07-17 (+8) \\\toprule
  
  \multirow{4}{*}{(c)} & CVE-2016-5195 & Race condition in mm/gup.c in the Linux kernel 2.x through 4.x before 4.8.3 allows local users to gain privileges by leveraging incorrect handling of a copy-on-write (COW) feature to write to a read-only memory mapping, as exploited in the wild in October 2016, aka ``Dirty COW."
& \begin{tabular}[t]{@{}l@{}}7.2 HIGH (v2.0) \\ 7.8 HIGH (v3.0)\end{tabular} & 2016-11-10 \\\cmidrule{2-5}
 & Linux & Serious Dirty COW bug leaves millions of Linux users vulnerable to attack: A vulnerability discovered in the ... http://tinyurl.com/zjdp268  & 0.97 & 2016-10-22 (+19)\\
& Linux OS & A critical vulnerability has been discovered in all versions of the Linux OS and is being exploited in the wild http://ift.tt/2es31Xc &  0.95 & 2016-10-25 (+16)\\
 & Linux COW & Serious vulnerability found in the Linux COW, may have persisted for a decade. http://www.bbc.co.uk/news/technology-37728010?ocid=socialflow\_twitter ... http://arstechnica.com/security/2016/10/most-serious-linux-privilege-escalation-bug-ever-is-under-active-exploit/ ... & 0.82 & 2016-10-21 (+20)  \\\toprule
 
 \multirow{4}{*}{(d)} & CVE-2016-7855 & Use-after-free vulnerability in Adobe Flash Player before 23.0.0.205 on Windows and OS X and before 11.2.202.643 on Linux allows remote attackers to execute arbitrary code via unspecified vectors, as exploited in the wild in October 2016. & \begin{tabular}[t]{@{}l@{}} 10.0 HIGH (v2.0)\\\,9.8 CRITICAL (v3.0)\end{tabular} & 2016-11-01 \\\cmidrule{2-5}
  & Flash & ICYMI Critical vulnerability found in Flash, being actively exploited. Patch Flash NOW https://www.grahamcluley.com/patch-flash/  & 0.97 & 2016-10-27 (+5) \\
 & Adobe & Adobe has released a Flash Player update to patch a critical vulnerability that malicious actors have been ex... http://bit.ly/2eaTxhO & 0.95 & 2016-10-26 (+6)\\
 & \begin{tabular}[t]{@{}l@{}} Adobe Flash \\ Player\end{tabular} & A critical vulnerability for Adobe Flash Player that allows an attacker to take control of the affected system. https://helpx.adobe.com/security/products/flash-player/apsb16-36.html ... & 0.80 & 2016-10-27 (+5) \\
 \bottomrule
\end{tabular}%
}
\caption{\label{tb:forecast_examples} Top 4 threats identified by our forecast model. Severity scores are generated by using threat severity classifier in \Cref{sec:predict}.}
\end{table*}

\Cref{tb:forecast_examples} presents top 4 forecast results from our model. We observe that our model can predict accurate severity level even 19 days ahead of the official published date in NVD (\Cref{tb:forecast_examples}(a), (c)).

\subsection{Predicting Real-World Exploits}
\label{sec:real_exploit}
In addition to comparing our forecasted severity scores against CVSS, as described above, we also explored several alternatives suggested by the security community to evaluate our methods: (1) Symantec's anti-virus (AV) signatures\footnote{\tiny \urlstyle{sf}\url{https://www.symantec.com/security-center/a-z}} and intrusion-protection (IPS) signatures,\footnote{\tiny \urlstyle{sf}\url{https://www.symantec.com/security_response/attacksignatures/}} in addition to (2) Exploit Database (EDB).\footnote{\tiny \urlstyle{sf}\url{https://www.exploit-db.com/}}

\citet{191006} suggested Symantec's AV and IPS signatures are the best available indicator for real exploitable threats in the wild. We follow their method of explicitly querying for CVE numbers from the descriptions of signatures to generate exploited threats ground truth. Exploit Database (EDB) is an archive of public exploits and software vulnerabilities. We query EDB for all threats that have been linked into NVD.\footnote{\tiny\urlstyle{sf}\url{http://cve.mitre.org/data/refs/refmap/source-EXPLOIT-DB.html}}  In total we gathered 134 CVEs verified by Symantec and EDB to be real exploits within the 1,409 CVEs used in our forecasting evaluation.

We evaluate the number of exploited threats identified within our top ranked CVEs. \Cref{tb:real_exp_res} presents our results. We observe that 7 of top 10 threats from our model were exploited in the wild. We also observe that for the actual CVSS v3.0 scores, only 1 out of the top 10 vulnerabilities was exploited.

\begin{table}[!htbp]
\resizebox{0.48\textwidth}{!}{%
\centering
\begin{tabular}{l|cc|cc|cc}
\toprule
\multirow{2}{*}{} & \multicolumn{2}{c|}{Top 10} & \multicolumn{2}{c|}{Top 50} & \multicolumn{2}{c}{Top 100} \\
& P & R & P & R & P & R \\\midrule
True CVSS & 10.0 & 0.7 & 16.0 & 6.0 & 16.0 & 11.9 \\
Volume model & 60.0 & 4.5 & 22.0 & 8.2 & 19.0 & 14.2 \\
Our model & 70.0 & 5.2 & 28.0 & 10.4 & 21.0 & 15.7 \\
\bottomrule
\end{tabular}
}
\caption{\label{tb:real_exp_res} Model performance against real-world exploited threats identified by Symantec and Exploit-DB. ``True CVSS" refers to ranking CVEs based on actual CVSS scores in NVD. This model is only for reference and can not be used in real practice, as we do not know true CVSS scores when forecasting.}
\end{table}

\subsection{Identifying Accounts that Post Reliable Warnings}

Finally we perform an analysis of the reliability of individual Twitter accounts.
We evaluate all accounts with more than 5 tweets exceeding 0.5 confidence score from our severity classifier. \Cref{tb:account_ana} presents our results. Accounts in our data whose warnings were found to have highest precision when compared against CVSS include ``@securityaffairs" and ``@EduardKovacs", which are known to post security related information, and both have more than 10k followers.

\begin{table}[!h]
\small
\centering
\begin{tabular}{l|cc}
\toprule
Account Name & \# Corr / \# Fcst & Acc. (\%)\\\midrule
jburnsconsult & 15 / 15 & 100 \\
securityaffairs & 10 / 10 & 100 \\
EduardKovacs & 6 / 6 & 100\\
cripperz &  5 / 5 & 100\\
cipherstorm & 4 / 5 & 80\\
\bottomrule
\end{tabular}
\caption{List of users with top accuracies on forecasting severe cybersecurity threats.}
\label{tb:account_ana}
\end{table}


\section{Related Work}

There is a long history of prior work on analyzing users' opinions online \cite{wiebe2004learning}, a large body of prior work has focused on sentiment analysis \cite{pang2002thumbs,rosenthal2015semeval}, e.g., determining whether a message is positive or negative.  In this paper we developed annotated corpora and classifiers to analyze users' opinions toward the severity of cybersecurity threats reported online, as far as we are aware this is the first work to explore this direction.

Forecasting real-world exploits is a topic of interest in the security community. 
For example, \citet{Bozorgi:2010:BHL:1835804.1835821} train SVM classifiers to rank the exploitability of threats. 
Several studies have also predicted CVSS scores from various sources including text descriptions in NVD \cite{8094415,Bullough:2017:PED:3041008.3041009}. 

Prior work has also explored a variety of forecasting methods that incorporate textual evidence \cite{smith2010text}, including the use of Twitter message content to forecast influenza rates \cite{paul2014twitter}, predicting the propagation of social media posts based on their content \cite{Tan2014TheEO} and forecasting election outcomes \cite{o2010tweets,swamy2017have}.


\section{Conclusion}

In this paper, we presented the first study of the connections between the severity of cybersecurity threats and language that is used to describe them online. We annotate a corpus of 6,000 tweets describing software vulnerabilities with authors' opinions toward their severity, and demonstrated that our corpus supports the development of automatic classifiers with high precision for this task. 
Furthermore, we demonstrate the value of analyzing users' opinions about the severity of threats reported online as an early indicator of important software vulnerabilities.  We presented a simple, yet effective method for linking software vulnerabilities reported in tweets to Common Vulnerabilities and Exposures (CVEs) in the National Vulnerability Database (NVD).  Using our predicted severity scores, we show that it is possible to achieve a Precision@50 of 0.86 when forecasting high severity vulnerabilities, significantly outperforming a baseline that is based on tweet volume.  Finally we showed how reports of severe vulnerabilities online are predictive of real-world exploits.

\section*{Acknowledgments}
We thank our anonymous reviewers for their valuable feedback. We also thank Tudor Dumitra\c{s} for helpful discussion on identifying real exploited threats.
Funding was provided by the the Office of the Director of National Intelligence (ODNI) and Intelligence Advanced Research Projects Activity (IARPA) via the Air Force Research Laboratory (AFRL) contract number FA8750-16-C0114, in addition to the Defense Advanced Research Projects Agency (DARPA) via the U.S. Army Research Office (ARO) and under Contract Number W911NF-17-C-0095, in addition to an Amazon Research Award and an NVIDIA GPU grant.  The content of the information in this document does not necessarily reflect the position or the policy of the Government, and no official endorsement should be inferred. The U.S. Government is authorized to reproduce and distribute reprints for government purposes notwithstanding any copyright notation here on.

\bibliography{ref}
\bibliographystyle{acl_natbib}


\appendix

\section{Linking Algorithm}

We describe our approach to match tweets with NVD records in full detail in \Cref{alg:matching}.

\begin{algorithm}[!h]
\small
  \caption{Linking tweets to NVD records.}
  \label{alg:matching}
  \begin{algorithmic}[1]
    \STATE // Linking
    \FOR{every tweet $t$}
    \IF{CVE number in tweet context or in url links}
    \STATE match CVEs to this tweet
    \ELSE
    \STATE query webpage contents to search for CVEs
    \ENDIF
    \ENDFOR
    \STATE 
    \STATE // Check linking results
    \STATE Keep tweets that matched to only one unique CVE to avoid ambiguities
    \STATE
    \STATE // Apply time constraints
    \STATE Select out tweets that are posted at least 5 days ahead of official NVD publication date (at most 365 days\footnotemark{})
  \end{algorithmic}
\end{algorithm}
\footnotetext{If a tweet or its associate urls explicitly contains a CVE number, then we ignore this maximum time range constraint.}

\section{Limitations of CVSS and Real-World Exploits Ground Truth}

In \Cref{sec:fore_cvss} - \Cref{sec:real_exploit}, we compare our forecast results with (1) CVSS ratings, and (2) real exploited threats identified by Symantec signatures and Exploit Database. Each of these sources of ground truth have limitations, which we discuss below.

CVSS ratings are widely used as standard indicators for risk measurement in practice.  However, one problem of CVSS ratings is that high severity threats do not necessarily lead to real-world exploits.  \citet{Allodi:2012:PAV:2382416.2382427} show that only a small portion (around 2\%) of reported vulnerabilities were found to be exploited in the wild.  
Furthermore, more than half of the threats in NVD are marked as HIGH or CRITICAL, causing a large burden on vendors to fix.\footnote{\tiny\urlstyle{sf}\url{https://www.riskbasedsecurity.com/2017/05/cvssv3-when-every-vulnerability-appears-to-be-high-priority/}} 
We also notice these CVSS scores are closely tied with specific categories of threats.  For example, 85.6\% of buffer errors are marked as HIGH or CRITICAL, while 72.5\% of information leaks were marked as MEDIUM or LOW. All these issues post challenges on how to prioritize real exploitable threats, with the goal of reducing false positives and false negatives simultaneously. Our work provides one such additional source of information for helping to prioritize threats.

The ground truth we use for real exploited threats is still an incomplete list. For example, Linux kernel vulnerabilities are less likely to appear in Symantec signatures, as Symantec does not have a security product for Linux. Identifying real exploited threats is a difficult task; to the best of our knowledge, there does not exist an easy-to-access list covering all exploited threats currently.

\begin{table*}[!h]
\centering
\footnotesize
\resizebox{\textwidth}{!}{%
\begin{tabular}{c|p{0.14\textwidth}p{0.16\textwidth}p{0.6\textwidth}|ll}
\toprule
 & CVE Num & Name Entity & Tweet & Our Score & Real Severity\\ \midrule
(a) & CVE-2017-4984 & EMC VNX1VNX2 OE &  threatmeter: Vuln: EMC VNX1/VNX2 OE for File CVE-2017-4984 Remote Code Execution Vulnerability http://ift.tt/2rWXQXa  & 0.01 & \begin{tabular}[t]{@{}l@{}} 10.0 HIGH (v2.0)\\\,9.8 CRITICAL (v3.0)\end{tabular}\\\midrule
\multirow{2}{*}{(b)} &\multirow{2}{*}{CVE-2016-1730} & iPhone & A newly discovered vulnerability may expose iPhone users to attack when using a Wi-Fi hotspot - via @InfosecurityMag http://owl.li/Xw3VO & 0.76 & \begin{tabular}[t]{@{}l@{}}5.8 MEDIUM (v2.0) \\5.4 MEDIUM (v3.0)\end{tabular}\\
& & iPhone& Apple iOS Flaw Enables Attacks via Hotspot: The vulnerability opens up iPhone users to a raft of problems, inc... http://bit.ly/1JqGtD9 & 0.45 \\
\bottomrule
\end{tabular}%
}
\caption{\label{tb:forecast_error} Some examples of forecast errors made by our model. (a) False negative examples: there is no clear language clue for demonstrating the severity of threats, experts are needed for threats of this kind. (b) False positive examples: there exist some signals captured by our model for being severe threats, but actual severity might be overestimated.}
\end{table*}

\section{Additional Analysis of Results}

In this section, we present further analyses of people's online behaviors when discussing cybersecurity threats on social media.

We find that the real severity of threats is predictable based on users' opinions online.
We observe several repeated patterns in how people describe severe threats. We summarize some of these patterns below:
\begin{itemize}[noitemsep]
\item  describing severity levels (see \Cref{sec:sub_adj}), such as ``critical", ``serious", ``highly'';
\item describing the number of users or devices affected, such as ``millions of $\langle$TARGET$\rangle$ devices", ``huge number of";
\item potential consequences, such as ``allows hackers to", ``could allow for remote code execution", ``malware";
\item alerts or warnings, such as ``please be aware", ``warning";
\item suggesting immediate actions, such as ``patch now".
\end{itemize}

\subsection{Usage of Subjective Adjectives}
\label{sec:sub_adj}

We notice people rely on adjectives for describing the level of severity for threats, rather than numerical scores. These subjective adjectives form our initial impressions on these threats.

We examine subjective adjectives people use for measuring threats. We run POS tagging to extract all tokens marked as {\ttfamily JJ}, {\ttfamily JJP}, and {\ttfamily JJS}. We then rank subjective adjectives in Subjectivity Lexicon (SUB) \cite{wilson-wiebe-hoffmann:2005:HLTEMNLP} by log-odds ratio of their occurrences in NVD descriptions for HIGH or CRITICAL threats versus MEDIUM or LOW threats. \Cref{tb:adj_res} presents top ranked subjective adjectives. We observe variants people are using for severe threats, e.g. ``serious", ``severe", ``malicious", etc.

\begin{table}[h!]
\centering
\small
\resizebox{0.48\textwidth}{!}{%
\begin{tabular}{lc|lc|lc}
\toprule
Adj. & Ratio & Adj. & Ratio & Adj. & Ratio\\\midrule
serious & 2.01 & aware & 1.61 & fast & 1.39 \\
pivotal & 1.95 & most & 1.61  & original & 1.39 \\
sure & 1.95 & vivid & 1.61 & able & 1.39 \\
free & 1.95 & accessible & 1.39 & blind & 1.39 \\
active & 1.79 & popular & 1.39 & arbitrary & 1.35 \\
intelligent & 1.79 & deep & 1.39 & high & 1.30 \\
static & 1.79 & black & 1.39 & incomplete & 1.25 \\
critical & 1.67 & top & 1.39 & malicious & 1.20 \\
severe & 1.61 & dangerous & 1.39 & wily & 1.10\\
great & 1.61 & wild & 1.39 & evil & 1.10\\
\bottomrule
\end{tabular}
}
\caption{Top ranked log-odds ratio of subjective adjectives describing severe threats (CVSS scores $\geq$ 7.0) versus non-severe threats (CVSS scores $<$ 7.0). Subjective adjectives are identified by using Subjectivity Lexicon (SUB) \cite{wilson-wiebe-hoffmann:2005:HLTEMNLP}.}
\label{tb:adj_res}
\end{table}

\subsection{Temporal Analysis}

We collect all CVEs having matched tweets posted at least 1 day ahead of the official NVD publication date, resulting in a set of 3,678 CVEs.  Within our dataset, 84.7\% of CVEs are reported within 60 days after the first disclosure on social media.  We observe a median of 5 days delay in our dataset, whereas some of threats have significant longer delays. 
For example, CVE-2016-2123\footnote{\tiny \urlstyle{sf}\url{https://nvd.nist.gov/vuln/detail/CVE-2016-2123}} (Overflow Remote Code Execution Vulnerability) first appears at Twitter on Dec. 19, 2016\footnote{\tiny \urlstyle{sf}\url{https://twitter.com/ryf_feed/status/810981102768758784}}, but is published in NVD on Nov. 1, 2018. 
It again shows the difficulty of threat evaluation and management.

\subsection{Error Analysis}
\label{sec:error_analysis}

We evaluate two types of errors with respect to forecasting high severity vulnerabilities: false positive and false negative examples.
We observe that some severe threats are difficult to predict based on contents in general, such as \Cref{tb:forecast_error}(a). There is no clear clue for estimating the severity level merely on tweet contents.  

We present another incorrect example extracted by our forecast system in \Cref{tb:forecast_error}(b). We notice tokens like ``expose users to attack", ``opens up to a raft of problems", etc. This threat does seem to be exploitable and harmful to a lot of users. However, experts mark it as of medium severity. It might be the case that the actual severity level of some threats are overestimated by some accounts.


\end{document}